\documentclass[10pt]{article}

\usepackage{arxiv}

\usepackage[utf8]{inputenc} % allow utf-8 input
\usepackage[T1]{fontenc}    % use 8-bit T1 fonts
\usepackage{hyperref}       % hyperlinks
\usepackage{url}            % simple URL typesetting
\usepackage{booktabs}       % professional-quality tables
\usepackage{amsfonts}       % blackboard math symbols
\usepackage{nicefrac}       % compact symbols for 1/2, etc.
\usepackage{microtype}      % microtypography
\usepackage{lipsum}		% Can be removed after putting your text content
\usepackage{graphicx}
\usepackage[numbers]{natbib}
\usepackage{doi}
\usepackage{multirow}

\usepackage{booktabs}
\usepackage{geometry}
\usepackage{colortbl}
\usepackage{xcolor}
\usepackage{pgfplots}
\pgfplotsset{compat=1.17}

\title{\textit{MicroVision}: An Open Dataset and Benchmark Models for Detecting Vulnerable Road Users and Micromobility Vehicles}

\author{ \href{https://orcid.org/0000-0001-6868-8364}{\includegraphics[scale=0.06]{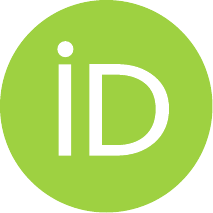}\hspace{1mm}Alexander~Rasch} \\
	Chalmers University of Technology\\
	Gothenburg, Sweden \\
	\texttt{alexander.rasch@chalmers.se} \\
	\And
	\href{https://orcid.org/0000-0002-1516-6930}{\includegraphics[scale=0.06]{orcid.pdf}\hspace{1mm}Rahul~Rajendra~Pai} \\
	Chalmers University of Technology\\
	Gothenburg, Sweden \\
	\texttt{rahul.pai@chalmers.se} \\
}

\hypersetup{
pdftitle={MicroVision: An Open Dataset and Benchmark Models for Detecting Vulnerable Road Users and Micromobility Vehicles},
pdfsubject={MicroVision: An Open Dataset and Benchmark Models for Detecting Vulnerable Road Users and Micromobility Vehicles},
pdfauthor={Alexander~Rasch, Rahul~Rajendra~Pai},
pdfkeywords={Micromobility, Open dataset, Traffic safety, Images, Bicycle, E-scooter, Object detection},
}

\begin{document}
\maketitle

\begin{abstract}
	Micromobility is a growing mode of transportation, raising new challenges for traffic safety and planning due to increased interactions in areas where vulnerable road users (VRUs) share the infrastructure with micromobility, including parked micromobility vehicles (MMVs). Approaches to support traffic safety and planning increasingly rely on detecting road users in images---a computer-vision task relying heavily on the quality of the images to train on. However, existing open image datasets for training such models lack focus and diversity in VRUs and MMVs, for instance, by categorizing both pedestrians and MMV riders as “person”, or by not including new MMVs like e-scooters. Furthermore, datasets are often captured from a car perspective and lack data from areas where only VRUs travel (sidewalks, cycle paths). To help close this gap, we introduce the MicroVision dataset: an open image dataset and annotations for training and evaluating models for detecting the most common VRUs (pedestrians, cyclists, e-scooterists) and stationary MMVs (bicycles, e-scooters), from a VRU perspective. The dataset, recorded in Gothenburg (Sweden), consists of more than 8,000 anonymized, full-HD images with more than 30,000 carefully annotated VRUs and MMVs, captured over an entire year and part of almost 2,000 unique interaction scenes. Along with the dataset, we provide first benchmark object-detection models based on state-of-the-art architectures, which achieved a mean average precision of up to 0.723 on an unseen test set. The dataset and model can support traffic safety to distinguish between different VRUs and MMVs, or help monitoring systems identify the use of micromobility. The dataset and model weights can be accessed at \url{https://doi.org/10.71870/eepz-jd52}.
\end{abstract}

% keywords can be removed

\section{Introduction} Micromobility refers to a mode of transport typically represented by compact and lightweight vehicles operating at lower speeds compared to more traditional modes of transport like passenger cars \citep{J3194_201911}. Micromobility has become a rising trend globally, particularly for personal transportation \citep{olabi2023}. Its rise has been explained by its compactness and cost-effectiveness---often making it a popular last-mile option---but also by improvements to health and sustainability, as it is a more environmentally friendly mode of transportation \citep{mcqueen2021,olabi2023}.

While the use of micromobility has increased over the last few years, so have interactions with other road users, as well as crashes \citep{yannis2024}. As most users of micromobility vehicles (MMVs) fall within the category of vulnerable road users (VRUs), they are less protected than users of motorized vehicles such as cars or trucks. Consequently, VRUs are more likely to be injured in collisions, particularly when they do not wear helmets or collide with heavier motorized vehicles. As a response to critical interactions with motorized vehicles, VRUs may also become psychologically discouraged from continuing to travel, which may have negative long-term consequences for society. Therefore, improving the safety of micromobility is paramount to further promote and sustain its use.

Following the safe systems approach \citep{khan2024}, initiatives to promote safe micromobility focus on road-user behavior, infrastructure, speed, and vehicle design. Initiatives focusing on behavior, for instance, study interactions between road users to understand what makes these interactions critical and to model behavior so that other initiatives can predict it and become more effective. For larger motorized vehicles, both active and passive safety systems have been developed to prevent crashes with VRUs or mitigate their consequences. However, for MMVs, such approaches have been scarce, partly due to tighter budget constraints and the limited space available for sensors and actuators.

A common challenge across all approaches is the need to accurately detect and distinguish micromobility vehicles and users in data. Such distinction is important because previous research has shown that e-scooterists, cyclists, and pedestrians have unique characteristics, such as differences in travel speed, route choice, and braking distance, for example during obstacle avoidance \citep{dozza2022,dozza2023,gilroy2022,kazemzadeh2023,pai2025}. For instance, an e-scooterist may appear from afar like a pedestrian, but travel at speeds comparable to a cyclist \citep{mura2022}. Detecting VRUs has become particularly important for image data captured by cameras, which are central to modern safety systems due to their relatively low cost, small size, and ability to capture the environment in a manner similar to human vision. However, developing object-detection models with high accuracy requires large amounts of diverse training data.

\section{Related Work} Previous work has provided multiple open traffic datasets that have proved useful to the research community developing models for detecting road users and vehicles \citep{alibeigi2023,geiger2013,sun2020}. However, these datasets have primarily been captured from a car perspective, thereby lacking data from cycle paths or sidewalks where only VRUs can travel. Furthermore, such datasets often miss finer distinctions between different types of micromobility, such as bicycles and e-scooters, or omit e-scooters entirely. E-scooters are known to perform differently compared to bicycles \citep{distefano2024,dozza2022}; therefore, distinguishing them from bicycles is important. In addition, detecting not only VRUs but also stationary MMVs has become increasingly important, as parked vehicles may pose obstacles to other road users.

Existing work on VRU detection has predominantly focused on pedestrians and cyclists. Most studies have developed annotated image datasets---some openly available---and trained deep-learning-based object-detection models. Model architectures have largely been based on \emph{convolutional neural networks} (CNNs), which can be broadly categorized into two main classes: (1) region-based (two-stage) detectors, such as R-CNN, Fast R-CNN, and Faster R-CNN, which first generate region proposals that are subsequently classified; and (2) single-shot (one-stage) detectors, such as YOLO and SSD, which directly predict object locations and classes \citep{garcia2021}. While two-stage detectors generally achieve higher accuracy at the cost of slower inference, single-stage detectors are often preferred for applications requiring faster inference, potentially at the expense of accuracy. With the advent of \emph{transformers} in computer vision, transformer-based detectors such as DETR, RT-DETR, and RF-DETR have gained attention for their end-to-end formulation and strong performance in complex scenes, albeit with higher computational demands.

Previous research on cyclist detection has predominantly relied on CNN-based detection models. \citet{garcia2021}, for example, developed an image dataset and detection model for cyclists and their orientations, concluding that region-based detectors yielded more accurate predictions, while single-shot detectors offered faster but less accurate results that could be sufficient when combined with object tracking. \citet{li2016} presented a public cyclist dataset from China containing more than 20,000 annotated instances and reported consistent detection performance across several then state-of-the-art architectures, including Fast R-CNN.

Due to the relatively recent appearance of e-scooters compared to bicycles on public roads, research on detecting e-scooters in images has been limited. \citet{apurv2021} were among the first to present a public image dataset including e-scooters in the United States, consisting of approximately 10,000 images from 83 interaction scenes. Along with the dataset, they introduced a MobileNetV2-based benchmark model to distinguish e-scooter riders from pedestrians. \citet{gilroy2022} further improved rider classification by training a model on web-sourced images of partially occluded e-scooter riders. \citet{chen2024} published a dataset of approximately 2,000 images collected in Charlottesville, USA, with over 11,000 annotated objects, and presented benchmark detection models using one-stage YOLO architectures. \citet{sabri2024} introduced another public dataset of stationary MMVs, based on web-sourced videos featuring bicycles, e-scooters, and skateboards, and proposed a YOLOX-based detector augmented with temporal features, achieving improved detection accuracy.

Few studies have addressed the direct detection of e-scooterists, defined as the combined entity of rider and vehicle, and most have focused instead on image-level classification to distinguish them from cyclists \citep{apurv2021,gilroy2022}. Additionally, many datasets have been collected over relatively short time spans, potentially missing seasonal variations in environmental conditions and road-user appearance. Most available datasets including e-scooterists have also been recorded in North America \citep{apurv2021,chen2024,sabri2024}. To support better generalization of detection models, additional datasets from other regions, including Europe, are needed \citep{alibeigi2023}.

In this study, we address the lack of data for micromobility safety by introducing \emph{MicroVision}, an open dataset accompanied by initial benchmark object-detection models. The dataset comprises over 8,000 high-resolution anonymized images with more than 30,000 annotated instances of VRUs and MMVs. Unlike existing car-centric datasets, the images are captured from the ego perspective of micromobility users, covering VRU infrastructure often absent from other datasets and spanning a full annual cycle in Gothenburg, Sweden. Furthermore, we introduce a state-aware classification scheme that explicitly distinguishes between active riders (e.g., e-scooterists) and stationary vehicles (e.g., parked e-scooters), a distinction critical for downstream safety tasks such as trajectory prediction and risk assessment. Finally, we provide benchmark models based on state-of-the-art object-detection architectures to support future research on micromobility detection, particularly for traffic-safety applications.

\section{Method}

\subsection{Data Collection}
\begin{figure}[h]
\centering
\includegraphics[width=0.9\linewidth]{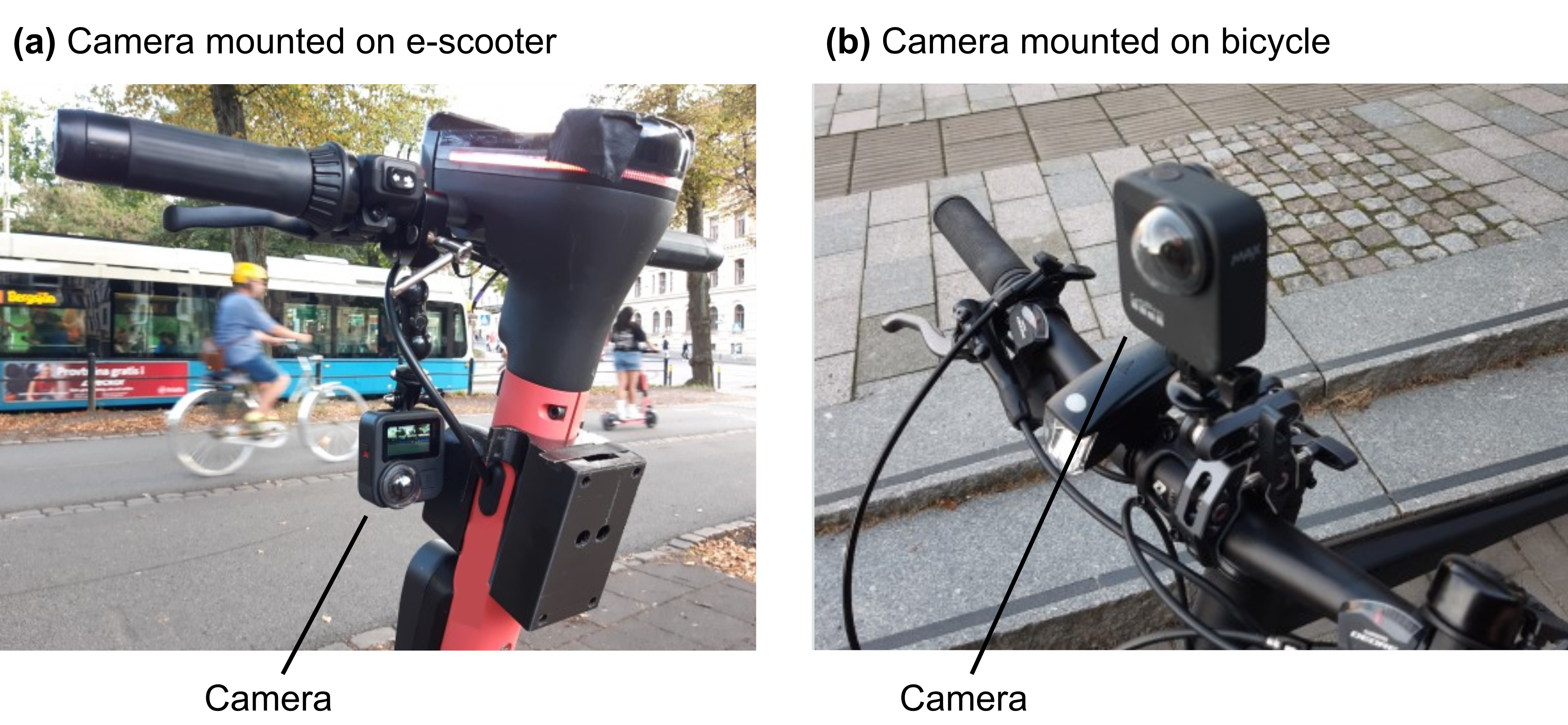}
\caption{Placement of the GoPro Max camera on the e-scooter (a) and bicycle (b) used for data collection.}
\label{fig:mounting}
\end{figure}

The images were derived from video footage systematically collected within the city of Gothenburg, Sweden. Data acquisition spanned a full annual cycle (from July 2024 to June 2025) and multiple times of day, thereby capturing a wide range of seasonal and lighting conditions to ensure substantial environmental and temporal diversity. Footage was captured from a micromobility user’s perspective by recording the surroundings while (1) riding an e-scooter, (2) riding a bicycle, and (3) walking. A GoPro Max camera was used for all recordings. For vehicle-based recordings, the camera was mounted on the handlebar at heights varying between 1.0 and 1.5 meters above the ground (see Fig.~\ref{fig:mounting}); while walking, the camera was handheld. Videos were recorded at a resolution of 1920 $\times$ 1080 pixels at 60 frames per second. The high frame rate ensured sharp captures of moving road users at close proximity. To minimize lens distortion and provide images suitable for standard object-detection architectures without intensive post-calibration, the camera was operated in a \emph{linear} lens mode.

The systematic recording of public spaces raises ethical concerns regarding the incidental collection of personal data without explicit individual consent. Requesting consent from each road user is impractical and may affect the realism of captured scenes. To address this, the data collector displayed a clearly visible sign during all recording sessions informing nearby road users about the ongoing data collection and providing contact information for inquiries or data-removal requests. Prior to public release, all data underwent anonymization to obscure identifiable personal information and comply with local data-protection regulations. Specifically, faces and vehicle license plates were blurred using the Precision Blur service provided by brighter AI, following prior work \citep{alibeigi2023}. The collection protocol and data provision were approved by the Swedish Ethical Review Authority (reference number 2024-00329-01).

\subsection{Frame Extraction}
From the recorded videos, relevant frames were extracted for annotation. The extraction aimed to maximize environmental diversity while maintaining a balanced distribution of object classes. Only a limited number of frames depicting the same object from different angles were selected. This process was facilitated by pre-annotating the full videos with the current best model (YOLO11-based, fine-tuned from public COCO-pretrained weights). To maintain consistent object identities across frames, we employed the BoT-SORT tracker \citep{aharon2022}. By tracking objects across frames and selecting frames sufficiently spaced in time, we ensured coverage of different viewpoints. Each candidate frame was then manually verified to confirm visual diversity and to exclude near-duplicate scenes.

\subsection{Annotations}
Selected frames were annotated with 2D rectangular bounding boxes for common VRUs and MMVs. Five classes were defined: (1) \emph{pedestrian}, (2) \emph{cyclist}, (3) \emph{e-scooterist}, (4) stationary \emph{bicycle}, and (5) stationary \emph{e-scooter}. Consistent with the National Highway Traffic Safety Administration definition, a pedestrian was defined as a person who is walking or sitting, provided they were not on a vehicle \citep{nhtsa2021}. For micromobility, we adopted a state-aware labeling strategy: a cyclist or e-scooterist was defined as the combination of the vehicle and at least one person traveling with it, including cases where the rider was stationary (e.g., waiting at a traffic light) or mounting the vehicle. Conversely, a person pushing a vehicle was labeled as two separate objects: a pedestrian and a vehicle. Stationary vehicles were annotated regardless of their orientation (upright or fallen).

Tricycles (e.g., bicycles with two front wheels) were included in the bicycle category in accordance with Swedish vehicle classification standards. Mopeds and large cargo cycles used for deliveries were excluded due to their rare appearance to avoid dataset imbalance. Smaller attachments (e.g., child seats or backpacks) were included within bounding boxes, while larger pushed or pulled objects (e.g., suitcases or child buggies) were excluded. Broken vehicles (e.g., missing wheels) were included in MMV categories. Following the COCO dataset conventions \citep{lin2014}, occluded objects were annotated with boxes covering the visible extent; in cases of severe occlusion (e.g., dense bicycle racks), only the closest and most distinguishable object was annotated.

\begin{figure}[h]
\centering
\includegraphics[width=\linewidth]{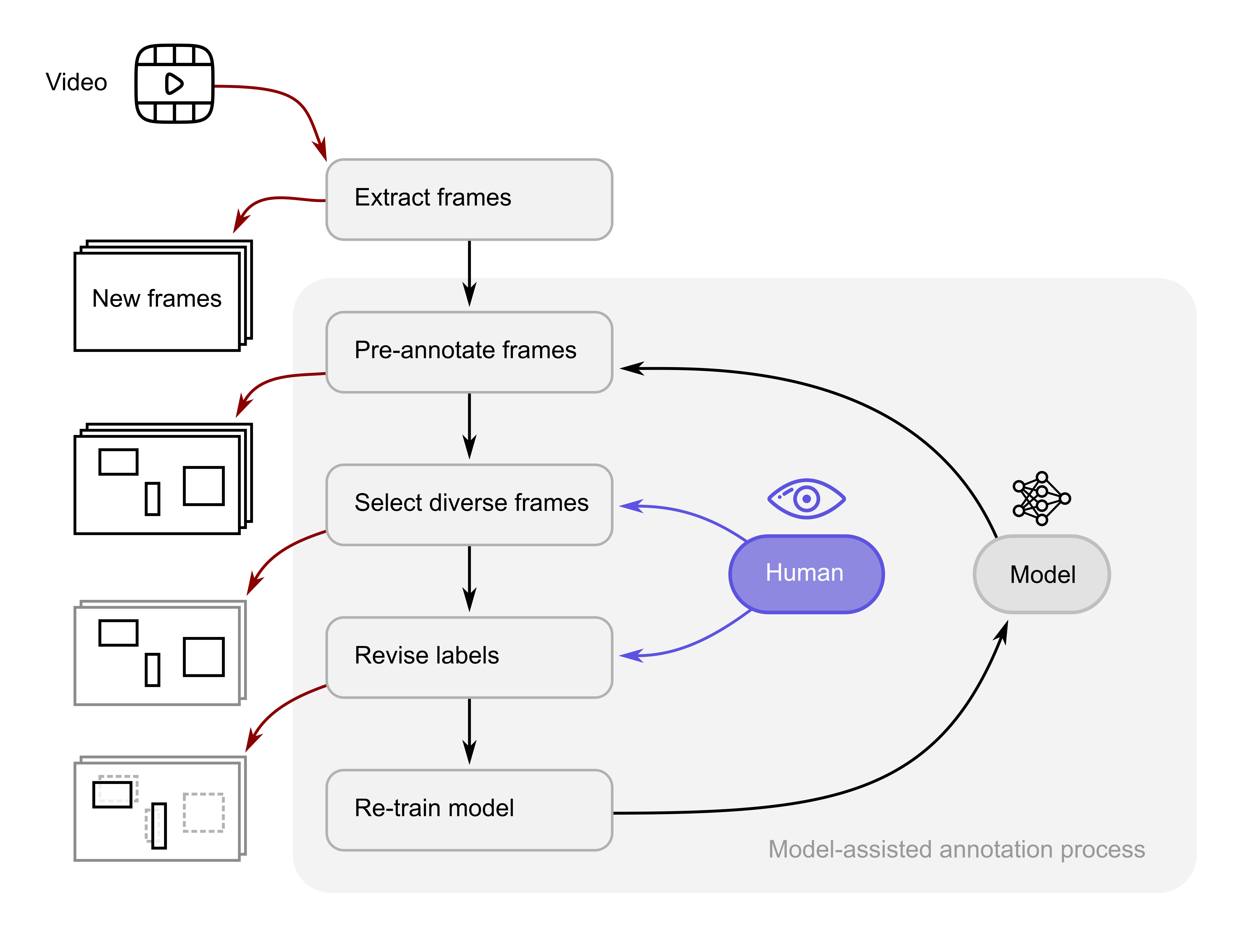}
\caption{Overview of the data processing pipeline, from raw videos to revised labels used for model training.}
\label{fig:pipeline}
\end{figure}

To maximize efficiency and consistency, a model-assisted semi-automated annotation workflow was employed (Fig.~\ref{fig:pipeline}). First, an object-detection model (YOLO7; \citet{wang2023}) was fine-tuned on a small set of open-source web images; \citep{fang2024}. This model was used to pre-annotate the first batch of images. Three human annotators then reviewed and corrected labels using Label Studio v1.13.1 \citep{LabelStudio}. The corrected data were used to fine-tune a new model for pre-annotating subsequent batches. This iterative predict-correct-retrain cycle was repeated until the entire dataset was processed, progressively reducing manual effort.

A final model-guided quality check was conducted to identify and correct potential human annotation errors. The best-performing model from the iterative loop was fine-tuned on the full dataset (90\%/10\% train/validation split) and used to predict labels for all images. Discrepancies between predictions and ground truth were manually reviewed, focusing on false negatives and false positives, and corrected as needed. This validation was repeated with varying splits until no labeling issues were observed.

To mitigate model bias toward object-dense scenes and reduce false positives in empty environments, a small number of background images (approximately 5\% of the dataset) containing none of the defined classes were added. These images were identified by the model as having no detections and verified manually.

\subsection{Inter-Annotator Agreement}
To assess reliability and consistency, an inter-annotator agreement analysis was conducted on a randomly selected subset of 258 images independently labeled by the three annotators. Pairwise agreement metrics were computed for all annotator pairs and averaged. For each pair, annotations were matched using the Hungarian algorithm to maximize total Intersection over Union (IoU) for bounding boxes of the same class \citep{kuhn1955}. Matches were considered valid only if IoU exceeded 0.5, consistent with common object-detection benchmarks such as Pascal VOC \citep{everingham2010}. Based on valid matches, \emph{box agreement} was computed as the average IoU to quantify spatial precision. \emph{Class agreement} was assessed using Cohen’s Kappa \citep{cohen1960}, accounting for chance agreement. Metrics were stratified by object class and object size to ensure robustness across road-user types and distances.

\subsection{Benchmark Object-Detection Models}
To provide initial performance benchmarks, three state-of-the-art object-detection architectures representing distinct paradigms were trained and evaluated: (1) one-stage YOLO version 11 (YOLO11; \citep{jocher2024}), (2) two-stage Faster R-CNN \citep{ren2015}, and (3) transformer-based RF-DETR \citep{robinson2025}. Publicly available pretrained weights were fine-tuned for all models. To focus on architectural comparison rather than hyperparameter optimization, a uniform training setup was used. All models were trained for 100 epochs with an effective batch size of 32 and an input resolution of 1280 pixels (for RF-DETR, the nearest compatible resolution of 1232 pixels was used). YOLO11 (largest available model variant ``X'') was trained using the Ultralytics package (v8.3.103). Faster R-CNN (largest variant ``X101-FPN'') was trained using the Detectron2 package (v0.6; \citep{wu2019}). RF-DETR (largest variant ``large'') was trained using the rfdetr package (v1.3.0) provided by Roboflow. Training was performed using up to four parallel NVIDIA A100 GPUs (80 GB VRAM each) on the Alvis compute cluster provided by the National Academic Infrastructure for Supercomputing in Sweden (NAISS).

To avoid data leakage, the dataset was split by \emph{scenes} rather than by individual frames. Scenes were defined as temporal sequences containing the same objects, identified using YOLO11 tracking outputs; transitions between scenes were detected when the last tracked object disappeared for more than 10 s before the next appeared. Scenes were stratified into training (80\%), validation (10\%), and test (10\%) sets.

Model performance was evaluated on the held-out test set using mean average precision (mAP), both per class and averaged across classes \citep{beitzel2009}. Following common practice, mAP was computed as the average over IoU thresholds from 0.5 to 0.95 in increments of 0.05 (mAP@0.5:0.95). In addition, performance was analyzed by object size following the COCO protocol \citep{lin2014}. Given the higher image resolution (1920 pixels wide) compared to COCO, size thresholds were scaled proportionally: \emph{small} (area $<$ 96\textsuperscript{2} px\textsuperscript{2}), \emph{medium} (96\textsuperscript{2} px\textsuperscript{2} $\le$ area $<$ 288\textsuperscript{2} px\textsuperscript{2}), and \emph{large} (area $\ge$ 288\textsuperscript{2} px\textsuperscript{2}).

\section{Results}

\subsection{The MicroVision Dataset}

\begin{figure}[h] \centering \includegraphics[width=\linewidth]{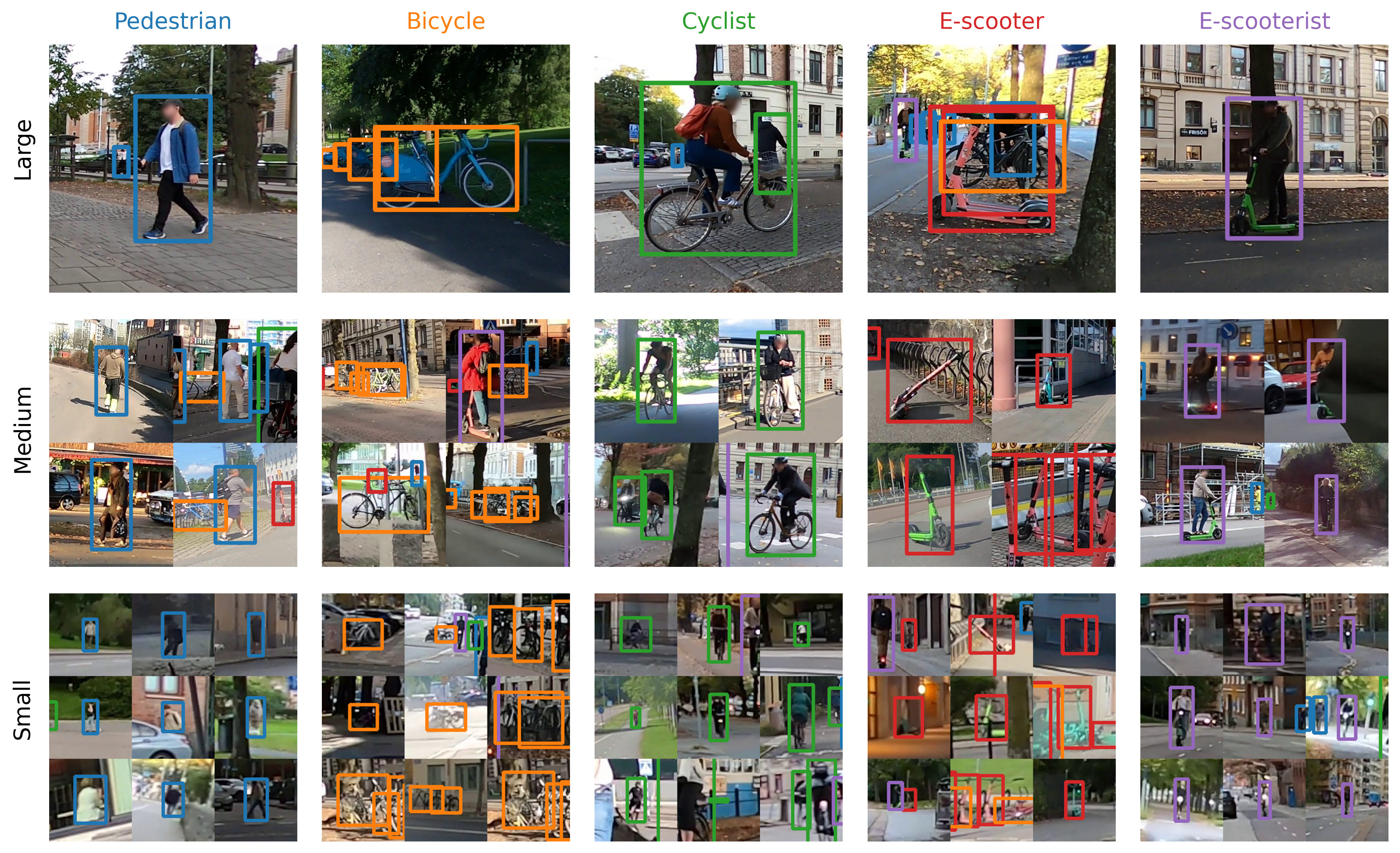} \caption{Example images and annotations from the MicroVision dataset for different object classes (columns) and object sizes (rows).} \label{fig:examples} \end{figure}

\begin{figure}[h] \centering \includegraphics[width=\linewidth]{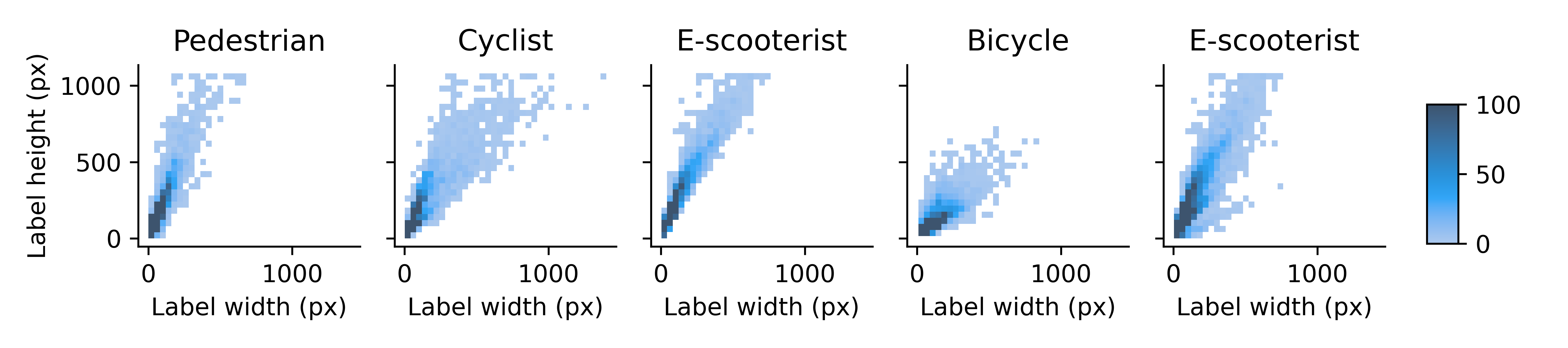} \caption{Bounding-box width and height distributions for the different object classes.} \label{fig:boxdist} \end{figure}

The MicroVision dataset contains 8,706 images, of which 594 are background images without any annotated objects. The remaining 8,113 images are annotated with a total of 34,866 objects using 2D bounding boxes. These include 17,032 pedestrians, 4,772 cyclists, 5,091 e-scooterists, 4,289 bicycles, and 3,682 e-scooters (see Fig.~\ref{fig:examples} for some examples). The images are organized into 1,984 unique scenes. Figure~\ref{fig:boxdist} shows the distributions of bounding-box widths and heights for all classes. Overall, the dataset consists of 67\% small, 25\% medium, and 8\% large objects. The comparatively large number of pedestrians, particularly small pedestrians, is a consequence of recording in dense urban environments, which increases background pedestrian frequency. While MMVs (bicycles and e-scooters) are predominantly represented by smaller and lower bounding boxes, VRU bounding boxes tend to be taller, as they combine both vehicle and rider.

The inter-annotator agreement analysis yielded an overall Cohen’s Kappa of 0.824 and an average Intersection over Union (IoU) of 0.887. Agreement correlated positively with object size, increasing from small objects (Kappa = 0.783; IoU = 0.866) to large objects (Kappa = 0.963; IoU = 0.964). Across classes, active road users (cyclists and e-scooterists) exhibited higher annotation consistency than stationary vehicles, as summarized in Table~\ref{tab:iaa}.

\begin{table}[h] 
\caption{Inter-annotator agreement for different object classes and object sizes (S = small, M = medium, L = large). Class agreement is reported using Cohen’s Kappa, and box agreement using mean Intersection over Union (IoU).} \label{tab:iaa}
\centering
\small
\setlength{\tabcolsep}{6pt}
\renewcommand{\arraystretch}{1.15}

\begin{tabular}{lcccccccc} 
\toprule
& \multicolumn{4}{c}{Class agreement (Cohen’s Kappa)} & \multicolumn{4}{c}{Box agreement (mean IoU)}\\
Class & S & M & L & All & S & M & L & All\\
\midrule
Pedestrian & 0.794 & 0.943 & 0.939 & 0.811 & 0.857 & 0.914 & 0.967 & 0.866\\
Bicycle & 0.676 & 0.837 & 1.000 & 0.761 & 0.872 & 0.891 & 0.955 & 0.886\\
Cyclist & 0.873 & 0.989 & 0.967 & 0.922 & 0.897 & 0.942 & 0.974 & 0.922\\
E-scooter & 0.670 & 0.837 & 0.897 & 0.728 & 0.847 & 0.904 & 0.968 & 0.873\\
E-scooterist & 0.877 & 0.987 & 0.978 & 0.938 & 0.911 & 0.953 & 0.957 & 0.937\\
All classes & 0.783 & 0.919 & 0.963 & 0.824 & 0.866 & 0.922 & 0.964 & 0.887\\
\bottomrule 
\end{tabular} 
\end{table}

\subsection{Object-Detection Benchmark}

Among the evaluated models, RF-DETR achieved the best overall performance on the held-out test set, with an overall mAP of 0.726, followed by YOLO11 (0.687) and Faster R-CNN (0.580), as reported in Table~\ref{tab:map}. While RF-DETR  excelled in overall performance for each object size and class, and in particular for medium and large objects, YOLO11 outperformed the other models on small pedestrians and cyclists, small and medium-sized e-scooters, as well as small e-scooterists, almost closing the performance gap to RF-DETR for small objects overall. Faster R-CNN performed consistently worse across all object classes and object sizes. Image anonymization resulted in negligible performance differences across all models (YOLO11: 0.690 original vs. 0.687 anonymized; Faster R-CNN: 0.579 original vs. 0.580 anonymized; RF-DETR: 0.731 original vs. 0.726 anonymized).

\begin{table}[h]
\caption{
Comparison of YOLO11, Faster R-CNN, and RF-DETR on the test set using COCO mAP@[0.5:0.95].
S, M, and L denote small, medium, and large objects, respectively.
Bold values indicate the best performance per class and object size.
}
\label{tab:map}
\centering
\small
\setlength{\tabcolsep}{6pt}
\renewcommand{\arraystretch}{1.15}

\begin{tabular}{llcccc}
\toprule
\textbf{Class} & \textbf{Model} & \textbf{S} & \textbf{M} & \textbf{L} & \textbf{All} \\
\midrule

\multirow{3}{*}{Pedestrian}
 & YOLO11     & \textbf{0.442} & 0.645 & 0.769 & 0.597 \\
 & Faster R-CNN & 0.279 & 0.559 & 0.726 & 0.499 \\
 & RF-DETR      & 0.438 & \textbf{0.673} & \textbf{0.869} & \textbf{0.629} \\

\midrule
\multirow{3}{*}{Bicycle}
 & YOLO11     & 0.110 & 0.373 & 0.724 & 0.518 \\
 & Faster R-CNN & 0.131 & 0.280 & 0.633 & 0.431 \\
 & RF-DETR      & \textbf{0.233} & \textbf{0.411} & \textbf{0.818} & \textbf{0.597} \\

\midrule
\multirow{3}{*}{Cyclist}
 & YOLO11     & \textbf{0.353} & 0.700 & 0.883 & 0.766 \\
 & Faster R-CNN & 0.093 & 0.570 & 0.818 & 0.669 \\
 & RF-DETR      & 0.332 & \textbf{0.725} & \textbf{0.932} & \textbf{0.813} \\

\midrule
\multirow{3}{*}{E-scooter}
 & YOLO11     & \textbf{0.232} & \textbf{0.609} & 0.837 & 0.692 \\
 & Faster R-CNN & 0.052 & 0.408 & 0.694 & 0.510 \\
 & RF-DETR      & 0.226 & 0.583 & \textbf{0.879} & \textbf{0.702} \\

\midrule
\multirow{3}{*}{E-scooterist}
 & YOLO11     & \textbf{0.574} & 0.787 & 0.927 & 0.865 \\
 & Faster R-CNN & 0.280 & 0.671 & 0.880 & 0.789 \\
 & RF-DETR      & 0.499 & \textbf{0.816} & \textbf{0.950} & \textbf{0.889} \\

\midrule
\multirow{3}{*}{All classes}
 & YOLO11     & 0.342 & 0.623 & 0.828 & 0.687 \\
 & Faster R-CNN & 0.167 & 0.497 & 0.750 & 0.580 \\
 & RF-DETR      & \textbf{0.346} & \textbf{0.641} & \textbf{0.890} & \textbf{0.726} \\

\bottomrule
\end{tabular}
\end{table}

\begin{figure}[h] \centering \includegraphics[width=\linewidth]{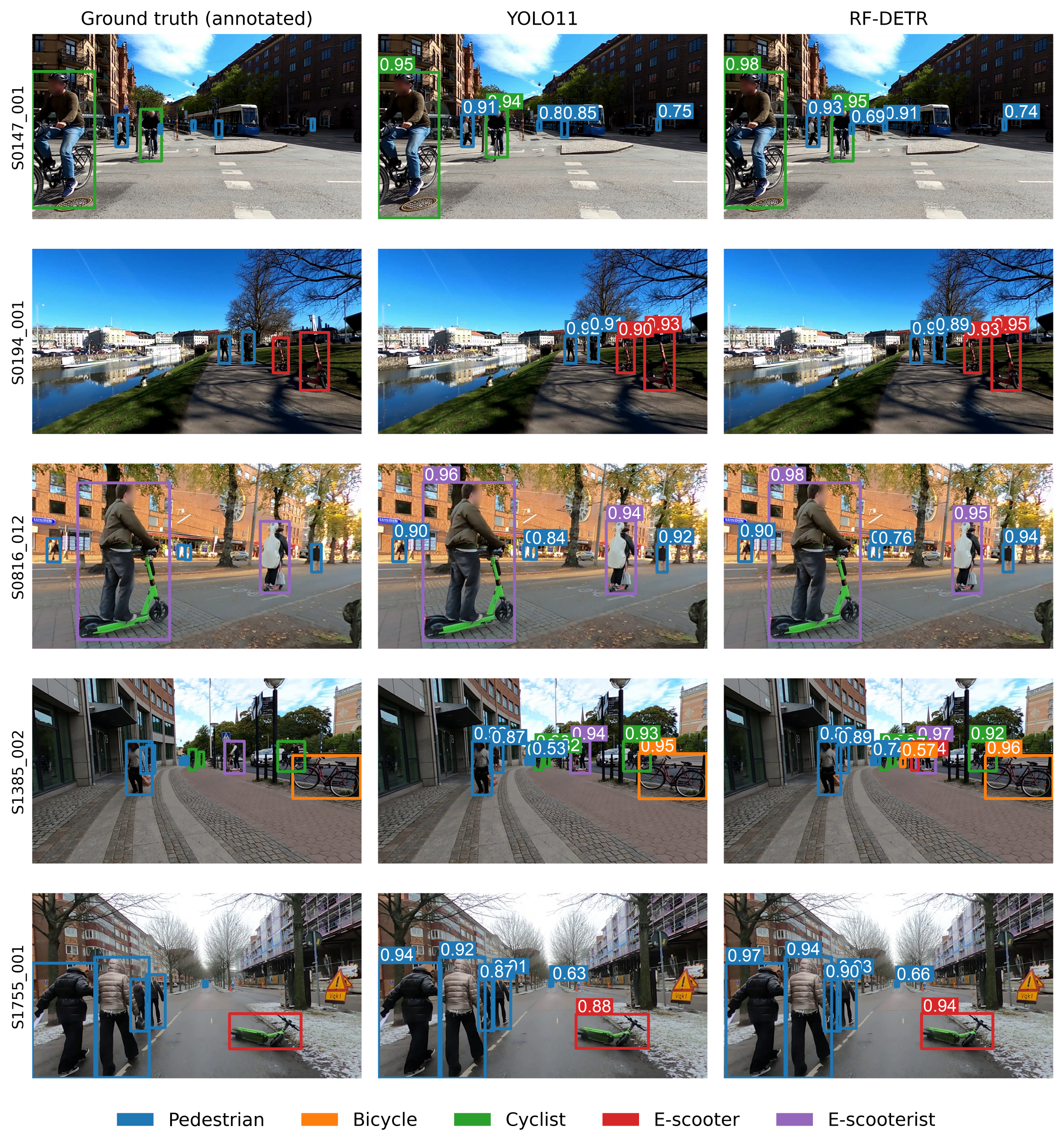} \caption{Successful example predictions on images from unseen test-set scenes for YOLO11 and RF-DETR, shown alongside ground-truth annotations. Numbers indicate predicted confidence scores (threshold = 0.5).} \label{fig:pred_good} \end{figure}

\begin{figure}[h] \centering \includegraphics[width=\linewidth]{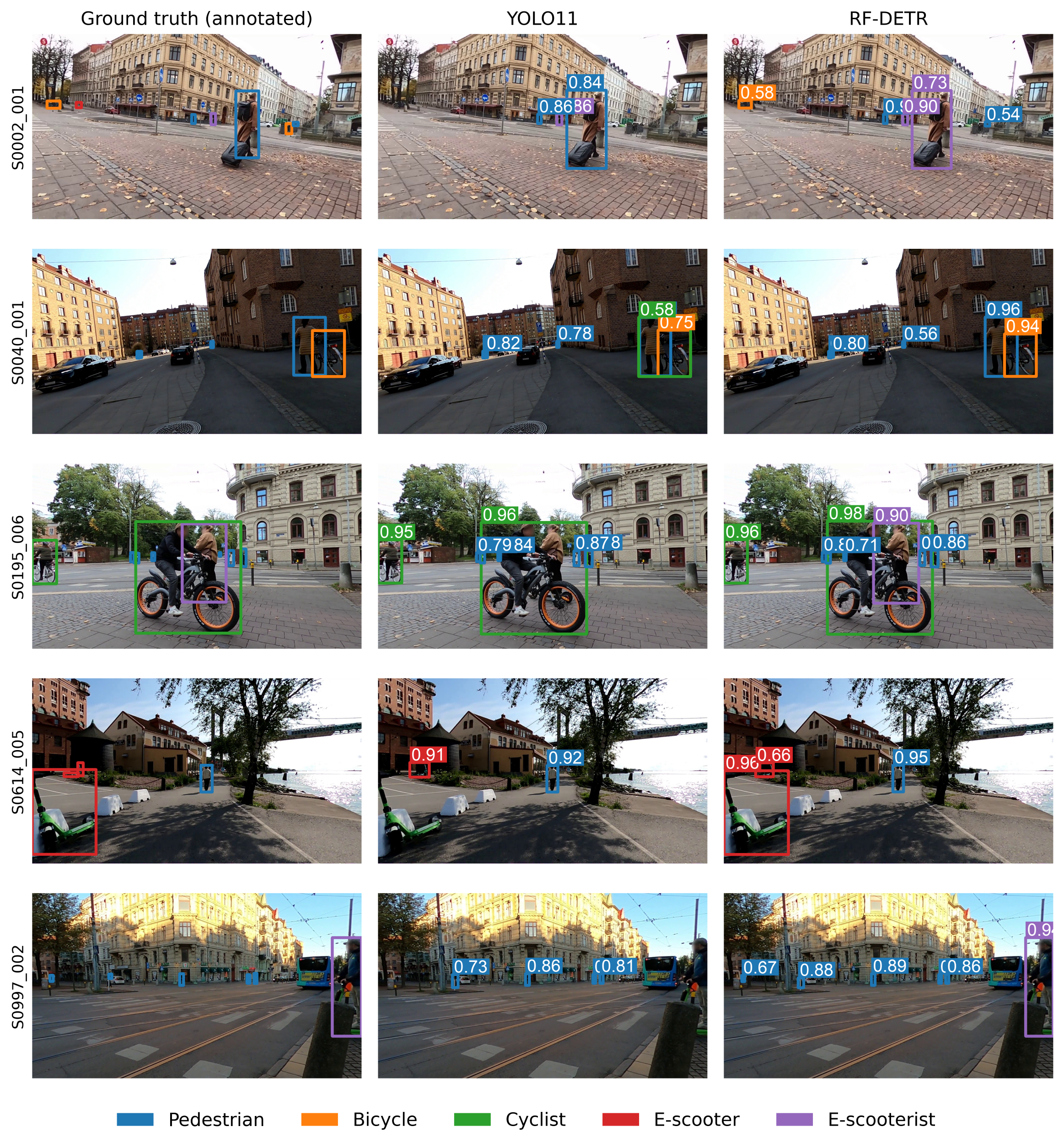} \caption{Representative examples of prediction errors from YOLO11 and RF-DETR, shown alongside ground-truth annotations.} \label{fig:pred_bad} \end{figure}

Figure~\ref{fig:pred_good} illustrates ground-truth annotations and predictions from the two best-performing models (YOLO11 and RF-DETR) on representative test images using a confidence threshold of 0.5. Figure~\ref{fig:pred_bad} highlights common failure cases. RF-DETR demonstrated superior performance for partially visible objects, such as occluded objects or objects near image boundaries. In some cases, both models incorrectly included adjacent objects (e.g., suitcases pulled by pedestrians) within bounding boxes; RF-DETR occasionally misclassified such instances. YOLO11, in contrast, correctly detected pedestrians pushing bicycles but sometimes produced redundant cyclist detections, indicating potential limitations in non-maximum suppression.

\section{Discussion}

\subsection{Dataset and Benchmark Models} The MicroVision dataset provides a specialized resource for understanding complex interactions in urban micromobility environments. While its scale of 8,706 images is smaller than many established car-centric datasets \citep{alibeigi2023,geiger2013,sun2020}, its scientific value lies in its qualitative uniqueness and technical specificity. Most existing open traffic datasets are recorded from the perspective of passenger cars and capture scenes primarily from the center of motorized traffic lanes. In contrast, the MicroVision dataset is captured directly from the perspective of micromobility users and pedestrians. This viewpoint shift is critical for developing safety systems intended for MMVs themselves, as it captures the specific visual context of sidewalks, cycle paths, and other shared urban spaces that are largely absent or underrepresented in car-centric data.

Furthermore, the dataset introduces a state-aware classification strategy that explicitly distinguishes between active riders (e.g., cyclists or e-scooterists) and stationary vehicles (e.g., bicycles or e-scooters). This distinction is essential for traffic safety and trajectory prediction, as a vehicle with a rider represents a dynamic entity with immediate kinetic potential and intent, whereas a parked vehicle constitutes a static obstacle. By providing high-resolution data collected over a full annual cycle, MicroVision ensures that these classifications are robust to environmental and seasonal variations typical of a Northern European urban setting. The large number of interaction scenes (nearly 2,000) further contributes to diversity in road-user appearance, viewing angles, and surrounding environments.

The initial benchmarking of state-of-the-art object-detection models demonstrated promising performance across architectures on unseen scenes. The transformer-based RF-DETR model outperformed CNN-based architectures (YOLO11 and Faster R-CNN) in overall performance, achieving an mAP of 0.723 and excelling particularly at detecting partially visible and occluded objects. This performance gap suggests that attention mechanisms may offer superior scene comprehension in dense urban environments. However, this improved accuracy comes at increased computational cost: the evaluated RF-DETR variant contains approximately 135 million parameters and requires longer inference times, making it less suitable for latency-critical real-time applications compared to more lightweight models such as YOLO11 (57 million parameters). All evaluated models struggled with reliably detecting stationary MMVs, largely due to dense clustering in parking zones where vehicles occlude one another. Such cases pose challenges not only for automated detection but also for consistent human annotation, as reflected in the inter-annotator agreement analysis.

Model performance further degraded for small and distant objects, such as far-away pedestrians, which is a well-known limitation in object detection. From a traffic-safety perspective, immediate collision risks are typically lower for such distant objects; however, applications requiring long-range perception may benefit from complementary techniques. Object tracking can mitigate missed detections over time, and Slicing Aided Hyper Inference (SAHI) can improve detection of small objects by performing inference on image subregions \citep{akyon2022}. Additional challenging cases included fallen or deformed (e.g., folded) MMVs, which were comparatively rare in the dataset and therefore more difficult for models to learn reliably.

\subsection{Implications and Applications} Within the transportation domain, the MicroVision dataset enables a wide range of applications. Traffic-safety research and development can leverage the dataset and benchmark models to build advanced driver or rider assistance systems for both motorized vehicles and MMVs, aiming to prevent collisions with VRUs and MMVs. Detection outputs can support downstream tasks such as trajectory prediction that account for distinct behavioral patterns, for example, the greater unpredictability often associated with e-scooters compared to bicycles \citep{distefano2024}.

Beyond real-time systems, the benchmark models can support retrospective traffic-safety research by enabling automated identification of road users in large-scale naturalistic video datasets \citep{beck2019,dozza2014,pai2025,schleinitz2017}. Such datasets are typically annotated manually, a process that is both time-consuming and error-prone. The MicroVision dataset and models may facilitate behavioral research by enabling efficient discovery of specific road-user interactions, appearances, and conflict scenarios.

Although primarily designed for safety-related research, the dataset may also benefit traffic management and urban planning. Automated detection and counting of VRUs can support infrastructure assessment, demand modeling, and policy evaluation \citep{peixoto2020,vacalebri2025}. However, the relatively low mounting height of the camera, corresponding to a road-user perspective, may limit the direct applicability of trained models to imagery captured from elevated viewpoints, such as building-mounted cameras.

\subsection{Limitations and Future Work} While the dataset captures a broad range of objects and environmental conditions, it is geographically limited to Gothenburg, Sweden. As a result, the generalization of trained models to regions with different infrastructure designs, traffic regulations, or vehicle types remains an open question. Moreover, most images were collected during daytime and under favorable weather conditions, reflecting periods of higher VRU activity. Future work should investigate whether additional data are required to improve generalization to nighttime or adverse weather conditions. Nonetheless, the dataset provides a strong foundation for future model-assisted annotation workflows, enabling efficient extension to new domains.

Regarding scope, the annotation effort focused on the most prevalent micromobility forms currently observed in Sweden. Less common VRUs and MMVs, such as low-speed mopeds or monowheels, were not included, nor were fine-grained distinctions within categories (e.g., bicycles with trailers). Future dataset expansions could address these gaps by extending the temporal and spatial coverage. Additionally, extending annotations beyond 2D bounding boxes to include instance segmentation could enable more precise pixel-level scene understanding and support tasks such as drivable-space estimation.

\section{Conclusion} We present an open image dataset with annotations and a first object-detection benchmark to advance the detection of vulnerable road users and micromobility vehicles. Comprising over 8,000 high-resolution images from nearly 2,000 unique interaction scenes and more than 30,000 annotations, the dataset addresses a critical gap by capturing the visual contexts of sidewalks and cycle paths and by employing a state-aware annotation strategy that distinguishes active riders from stationary vehicles.

Benchmarking state-of-the-art object-detection architectures shows that transformer-based models such as RF-DETR achieve superior accuracy in complex and occluded scenes, while efficient single-stage detectors like YOLO11 offer a favorable trade-off for real-time applications. Together, the dataset and benchmark models provide a foundation for next-generation traffic systems that can improve the safety and comfort of micromobility users, both through real-time assistance systems and by enabling partial automation of video-based behavioral research. By releasing these resources openly, we aim to support the development of models that generalize across the diverse and evolving landscape of urban transportation with a focus on micromobility.

\section*{Acknowledgments} The authors thank Shiyi Qiu, Mahin Garg, and Anton Broman (Chalmers University of Technology) for their assistance with data processing and annotation, and Marco Dozza (Chalmers University of Technology) for valuable discussions and funding acquisition. The authors also thank the Chalmers Data Office for support with the practical and administrative aspects of data set preparation and publication. 

The computations were enabled by resources provided by the National Academic Infrastructure for Supercomputing in Sweden (NAISS), partially funded by the Swedish Research Council through grant agreement no. 2022-06725. 

This work was carried out within the \emph{MicroVision} project, funded by Vinnova (Sweden’s innovation agency), the Swedish Energy Agency, and Formas (the Swedish Research Council for Sustainable Development) through the DriveSweden program (reference number 2023-01047).

\section*{Data Availability}
The MicroVision dataset and benchmark model weights are publicly available at \url{https://doi.org/10.71870/eepz-jd52}, hosted on the Data Organisation and Information System (DORIS) by the Swedish National Data Service (SND). Code to reproduce the analyses and data-processing pipeline is available at \url{https://github.com/microlab-chalmers/microvision}.

\section*{Declarations}
\subsection*{Conflict of Interest}
The authors declare no conflict of interest.

\bibliographystyle{unsrtnat}%{unsrtnat}
\bibliography{references}  %%% Uncomment this line and comment out the ``thebibliography'' section below to use the external .bib file (using bibtex) .





% \end{thebibliography}

\end{document}